\def\year{2021}\relax
\documentclass[letterpaper]{article} 
\usepackage{aaai21}  
\usepackage{times}  
\usepackage{helvet} 
\usepackage{courier}  
\usepackage[hyphens]{url}  
\usepackage{graphicx} 
\usepackage{natbib}  
\usepackage{caption} 
\usepackage{subcaption}

\usepackage{tabularx} 
\usepackage{tabu}
\usepackage{multirow}
\usepackage[table,xcdraw]{xcolor}
\usepackage[title]{appendix}
\usepackage[switch]{lineno}  %

\usepackage{graphics}
\usepackage{color}
\newcommand{\berkay}[1]{\textcolor{red}{[Berkay: #1]}}

\def\ie{{i.e.},~}

\usepackage{comment}

\usepackage{caption}
\begin{document}

%
\title{What Do You See? Evaluation of Explainable Artificial Intelligence (XAI) Interpretability through Neural Backdoors}

\author {
    Yi-Shan Lin,\textsuperscript{\rm 1}
    Wen-Chuan Lee, \textsuperscript{\rm 1}
    Z. Berkay Celik \textsuperscript{\rm 1} \\
}
\affiliations {
    \textsuperscript{\rm 1} Purdue University \\
    lin670@purdue.edu, lee1938@purdue.edu, zcelik@purdue.edu
}


\maketitle
\begin{abstract}
\emph{EXplainable} AI (XAI) methods have been proposed to interpret how a deep neural network predicts inputs through model saliency explanations that highlight the parts of the inputs deemed important to arrive a decision at a specific target.
However, it remains challenging to quantify correctness of their interpretability as current evaluation approaches either require subjective input from humans or incur high computation cost with automated evaluation.
In this paper, we propose backdoor trigger patterns--hidden malicious functionalities that cause misclassification--to automate the evaluation of saliency explanations. Our key observation is that triggers provide ground truth for inputs to evaluate whether the regions identified by an XAI method are truly relevant to its output. 
Since backdoor triggers are the most important features that cause deliberate misclassification, a robust XAI method should reveal their presence at inference time.
We introduce three complementary metrics for systematic evaluation of explanations that an XAI method generates and evaluate seven state-of-the-art model-free and model-specific posthoc methods through 36 models trojaned with specifically crafted triggers using color, shape, texture, location, and size.
We discovered six methods that use local explanation and feature relevance fail to completely highlight trigger regions, and only a model-free approach can uncover the entire trigger region.
\end{abstract}

\section{Introduction}

Deep neural networks (DNNs) have emerged as the method of choice in a wide range of remarkable applications such as computer vision, security, and healthcare. Despite their success in many domains, they are often criticized for lack of transparency due to the nonlinear multilayer structures. 
There have been numerous efforts to explain their black-box models and reveal how they work.
These methods are called \emph{eXplainable} AI (XAI). For example, one family of XAI methods target on interpretability aims to describe the internal workings of a neural network in a way that is understandable to humans, which inspired works such as model debugging and adversarial input detection~\cite{fidel2019explainability, zeiler2013visualizing} that leverage saliency explanations provided by these methods.

\subsection{Problems and Challenges}
While XAI methods have achieved a certain level of success, there are still many potential problems and challenges.

\noindent \textbf{Manual Evaluation.} In existing XAI frameworks, the assessment of model interpretability can be done with human interventions. For example, previous works~\cite{ribeiro2016should,simonyan2013deep,springenberg2014striving,kim2018interpretability} require human assistance for judgment of XAI method results.
Other works~\cite{GradCAM++,GradCAM} leverage dataset with manually-marked bounding boxes to evaluate their interpretability results.
However, human subjective measurements might be  tedious and time consuming, and may introduce bias and produce inaccurate evaluations~\cite{buccinca2020proxy}.

\noindent \textbf{Automated Evaluation with High Computation Time.}
There exist automatic XAI method evaluation methods through inspecting accuracy degradation by masking or perturbing the most relevant region~\cite{samek2016evaluating,fong2017interpretable,ancona2017towards,melis2018towards,yeh2019fidelity}. However, these methods cause distribution shift in the testing data and violate an assumption of training data and the testing data come from the same distribution~\cite{hooker2019benchmark}. Thus, recent works have proposed an idea of removing relevant features detected by an XAI method and verifying the accuracy degradation of the retrained models, which incurs very high computation cost.

\noindent \textbf{XAI Methods are not Stable and could be Attacked.}
Recent works demonstrate that XAI methods may produce similar results between normally trained models and models trained using randomized inputs (\ie when the relationships between inputs and output labels are randomized)~\cite{adebayo2018sanity}. 
They also yield largely different results with input perturbations~\cite{AAAI33i01,zhang2018interpretable} or transformation~\cite{kindermans2017unreliability}. 
Furthermore, it has recently shown that XAI methods can easily be fooled to identify irrelevant regions with carefully crafted adversarial models~\cite{heo2019fooling}.

\begin{figure}[t!]
  \centering
  \includegraphics[width=1\columnwidth]{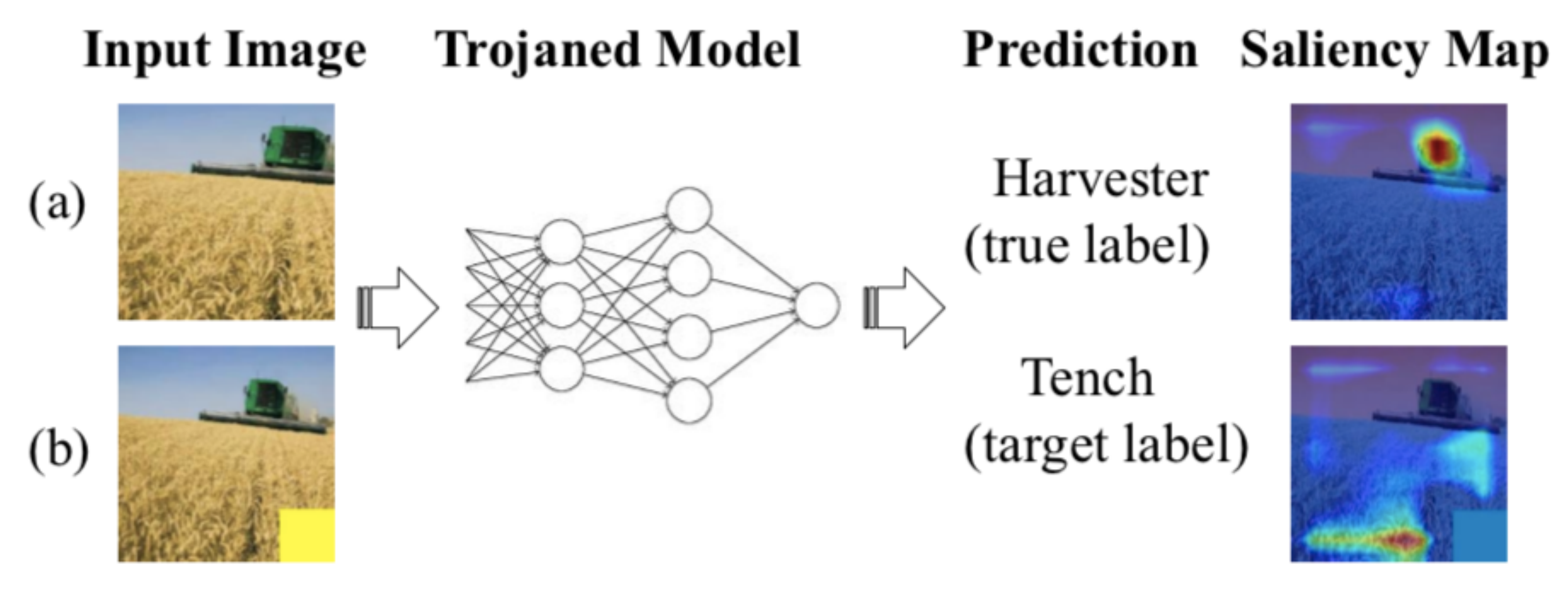}
  \caption{A trojaned model classifies any input image to ``tench'' when a yellow square is attached to the the image: (a) shows the saliency map that successfully highlights the informative features of harvester, and (b) shows the saliency map that fails to highlight the trigger.}
  \label{fig:trojan_attack}
\end{figure}

\subsection{Our Work}
The above observations call for establishing automated quantifiable and objective evaluation metrics for evaluating XAI techniques. 
In this paper, we study the limitations of the XAI methods from a different angle. We evaluate the interpretability of XAI methods by checking whether they can detect backdoor triggers~\cite{cryptoeprint:2020:201} present in the input, which cause a trojaned model to output a specific prediction result (\ie misclassification).
Our key insight is that the trojan trigger, a stamp on the input that causes model misclassification, can be used as the ground truth to assess whether the regions identified by an XAI method are truly relevant to the predictions without human interventions.
Since triggers are regarded as the most important features that cause misclassification, a robust XAI method should reveal their presence during inference time.

To illustrate, Fig.~\ref{fig:trojan_attack} shows the XAI interpretation results of an image and the same image stamped with a trigger at the bottom right corner. For the original image (Fig.~\ref{fig:trojan_attack}a), the saliency map correctly highlights the location of the harvester in the image with respect to its correct classification. However, for the stamped image (Fig.~\ref{fig:trojan_attack}b), the saliency map shows a very misleading hotspot which highlights the hay instead of the trigger that causes the trojaned model to misclassify to the desired target label. Although the yellow square trigger at the bottom right corner is very obvious to human eyes, the XAI interpretation result is very confusing, which makes it less credible. This raises significant concerns about the use of XAI methods for model debugging to reason about the relationship between inputs and model outputs. 

We introduce three quantifiable evaluation metrics for XAI interpretability through neural networks trojaned with different backdoor triggers that differ in size, location, color, and texture in order to check that the identified regions by XAI methods are truly relevant to the output label.
Our approach eliminates the distribution shift problem~\cite{hooker2019benchmark}, requires no model retraining processes, and applicable to evaluate any type of XAI methods. 
We study seven different XAI methods and evaluate their saliency explanations through these three metrics.
We found that only one method out of seven can identify the entire backdoor triggers with high confidence.
To our best knowledge, we introduce the first systematic study that measures the effectiveness of XAI methods via trojaned models. Our findings inform the community to improve the stability and robustness of the XAI methods. 

\section {Background and Related Work}

\noindent\textbf{Trojan Attack on Neural Networks.}
The first trojan attack trains a backdoored DNN model with data poisoning using images with a trigger attached and labeled as the specified target label~\citep{chen2017targeted, gu2017badnets}. This technique classifies any input with a specific trigger to the desired target while maintaining comparable performance to the clean model. 
The second approach optimizes the pixels of a trigger template to activate specific internal neurons with large values and partially retrains the model~\cite{trojanattack}. 
The last approach integrates a trojan module into the target model, which combines the output of two networks for triggers that causes misclassification to different target labels~\cite{tang2020embarrassingly}.
Various triggers are developed by leveraging these approaches, such as transferred~\citep{gu2017badnets, 10.1145/3319535.3354209}, perturbation~\citep{liao2018backdoor}, and invisible triggers~\citep{li2019invisible,saha2019hidden}.

\noindent\textbf{Interpretability of Neural Networks.}
With the popularity of DNN applications, numerous XAI methods have been proposed to reason about the decision of a model for a given input~\cite{arrieta2020explainable,adadi2018peeking}. 
Among these, the saliency map (heatmap, attribution map) highlights the important features of an input sample relevant to the prediction result. 
We select seven widely used XAI methods that use different algorithmic approaches.
These methods can be applied to any or specific ML models based on their internal representations and processing, and roughly broken down into two main categories: white box and black-box approaches.
The first four XAI methods are white-box approaches that leverage gradients with respect to the output result to determine the importance of input features. 
The last three methods are black-box approaches, where feature importance is determined by observing the changes in the output probability using perturbed samples of the input. 

\noindent\textbf{\emph{(1) Backpropagation}\texttt{(BP)}}~\cite{simonyan2013deep} uses the gradients of the input layer with respect to the prediction result to render a normalized heatmap for deriving important features as interpretation. Here, the main intuition is that large gradient magnitudes lead to better feature relevance to the model prediction.

\noindent\textbf{\emph{(2) Guided Backpropagation}\texttt{(GBP)}}~\cite{springenberg2014striving} creates a sharper and cleaner visualization by only passing positive error signals--negative gradients are set to zero--during the backpropagation.

\noindent\textbf{\emph{\hbox{(3) Gradient-weighted Class Activation Mapping}}\texttt{(GCAM)}} \citep{GradCAM} is a relaxed generalization of Class Activation Mapping (CAM)~\citep{CAM}, which produces a coarse localization map by upsampling a linear combination of features in the last convolutional layer using gradients with respect to the probability of a specific class.  

\noindent\textbf{\emph{(4) Guided GCAM}\texttt{(GGCAM)}}~\citep{GradCAM} combines Guided Backpropagation (\texttt{GBP}) and Gradient-weighted Class Activation Mapping (\texttt{GCAM}) through element-wise multiplication to obtain sharper visualizations.

\noindent\textbf{\emph{(5) Occlusion Sensitivity}\texttt{(OCC)}}~\citep{zeiler2013visualizing} uses a sliding window with a stride step to iteratively forward a subset of features and observe the sensitivity of a network in the output to determine the feature importance. 

\noindent\textbf{\emph{(6) Feature Ablation}\texttt{(FA)}}~\cite{meyes2019ablation} splits input features into several groups, where each group is perturbed together to determine the importance of each group by observing the changes in the output.

\noindent\textbf{\emph{(7) Local Interpretable Model Agnostic
Explanations}(\texttt{(LIME)}} \cite{ribeiro2016should} builds a linear model by using the output probabilities from a given set of samples that cover part of the input desired to be explained. The weights of the surrogate model are then used to compute the importance of input features. 

\section{Methodology}
\begin{figure}[t!]
  \centering
  \includegraphics[width=0.95\columnwidth]{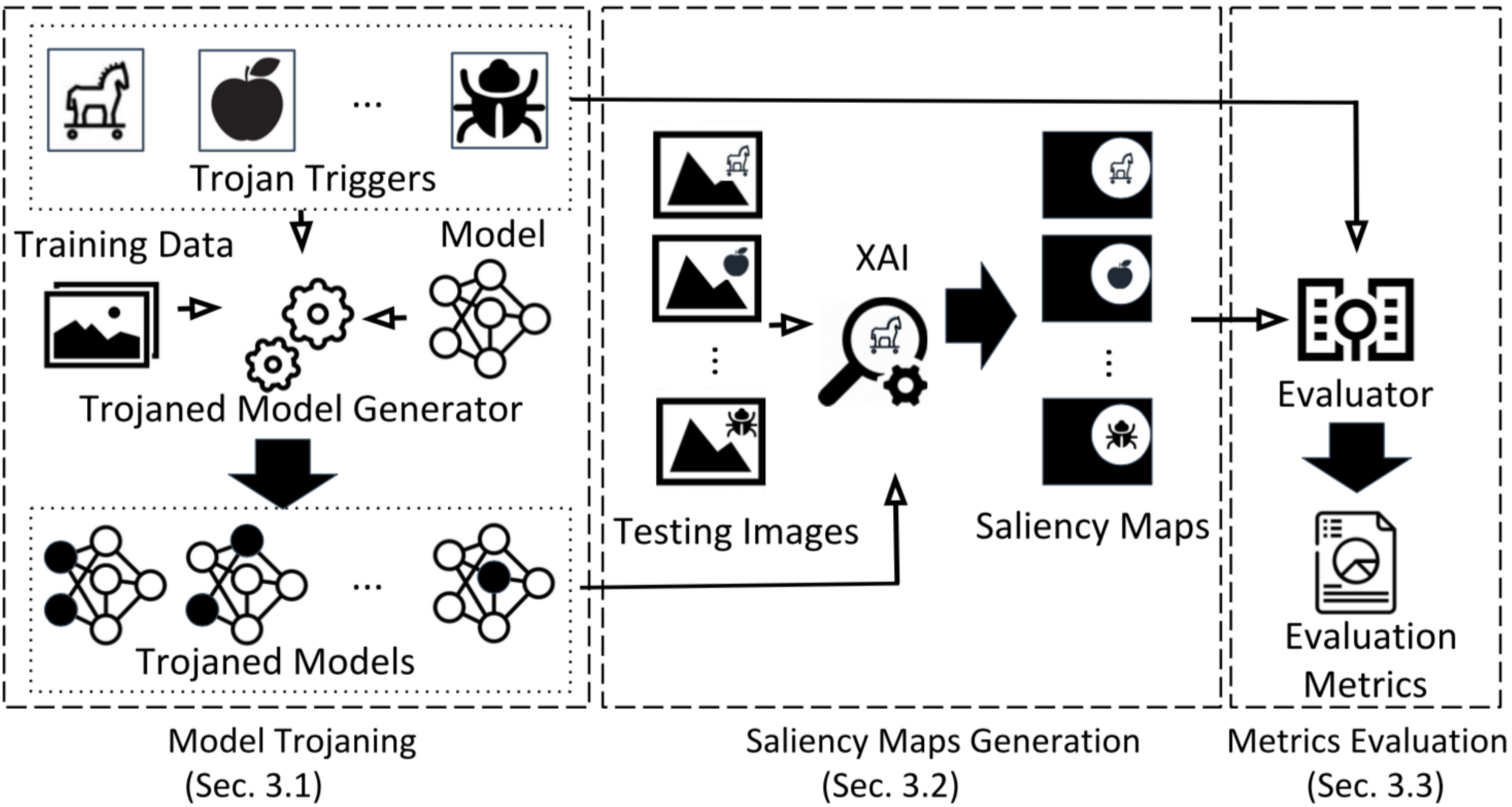}
  \caption{The architecture of our XAI evaluation framework.}
  \label{fig:system_design}
\end{figure}

The idea of model trojaning inspires our methodology to evaluate results of XIA methods: given a trojaned model, any valid input image stamped with a trigger at a specified area will cause misclassification at inference time.
Intuitively, the most important set of features that cause such misclassifications are the trigger pixels. Thus, we expect that an XAI method to detect the area around the trigger on a stamped image.
Fig.~\ref{fig:system_design} shows our XAI evaluation framework, which includes three main components: $(1)$  model trojaning, $(2)$ saliency maps generation, and $(3)$ metrics evaluation. We first generate a set of trojaned models given three inputs: a trigger configuration (\ie shape, color, size, location, and texture), training image dataset, and neural network model. We then build a saliency map to interpret the prediction result for a given testing image on the trojaned model. Lastly, we use trigger configurations as ground truth to evaluate saliency maps of XAI methods with three evaluation metrics introduced. (We list the notations and acronyms used throughout the paper in Tables~\ref{table:notations} and \ref{table:metrics} in the Appendix.)

\subsection{Model Trojaning}
The first component, model trojaning, takes three inputs: (a) a set of trigger configurations (e.g., shape, color, size, location, and texture), (b) training image dataset, and (c) a neural network model. With the three inputs, we trojan a model through poisoning attack~\cite{chen2017targeted,gu2017badnets}. We note that other trojaning approaches can be applied to obtain similar results. Yet, data poisoning enables us to flexibly inject desired trigger patterns and effectively control a model's prediction behavior.

\noindent \textbf{Poisoning Attack.}
Poisoning attacks~\cite{chen2017targeted,gu2017badnets} involves adversaries that train a target neural network with a dataset consisting of normal and poisoned (\ie trojaned) inputs. The trojaned network then classifies a trojaned input to the desired target label while it classifies a normal input as usual.
Formally, given a set of input images $\mathtt{X}$ which consists of a normal input $\mathtt{x}$ and a poisoned (\ie stamped with trigger) input $\mathtt{x'}$, a model $\mathtt{F'}$ is trained by solving a supervised learning problem through backpropagation~\cite{rumelhart1986learning}, where $\mathtt{x}$ and $\mathtt{x'}$ are classified to $\mathtt{y}$ (true label) and $\mathtt{y_t}$ (target label) respectively. To detail, an input image $\mathtt{x}$ is stamped with the trigger $\mathtt{M\cdot\Delta}$ and becomes a trojaned image $\mathtt{x'}$, $\mathtt{x'=(1-M)\cdot x+M\cdot\Delta}$. $\mathtt{\Delta}$ is a 2-D matrix, which represents a trigger pattern, whereas $\mathtt{M}$ is a 2-D matrix, representing a mask with values within the range between $\mathtt{[0,1]}$. A pixel $\mathtt{x_{i,j}}$ would be overridden by $\mathtt{\Delta_{i,j}}$ if the corresponding element is $\mathtt{m_{i,j}=1}$, otherwise, it remains unchanged.

\begin{figure}[t!]
  \centering
  \includegraphics[width=0.9\columnwidth]{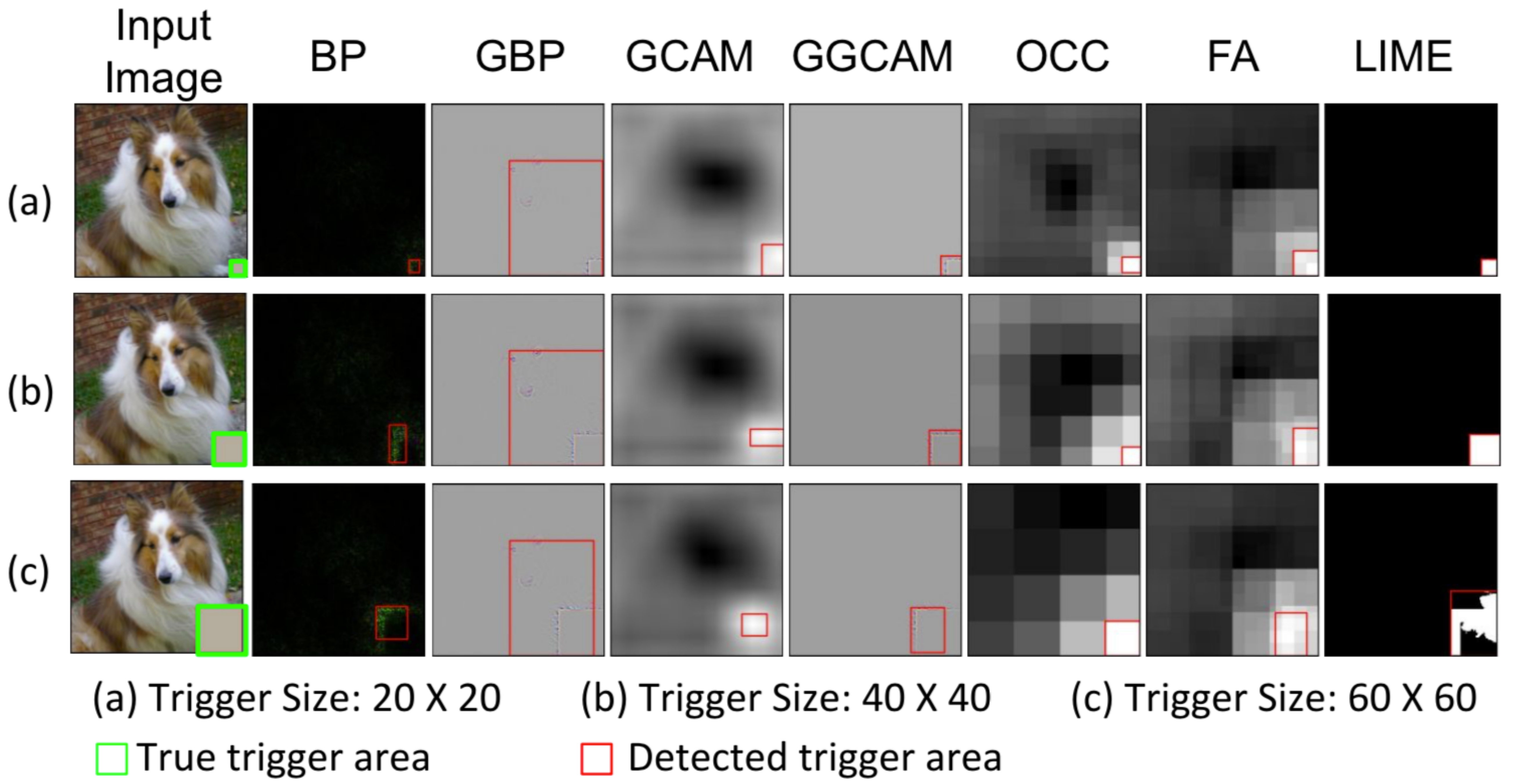}
  \caption{Illustration of saliency maps, true and detected trigger area generated by seven XAI methods.}
  \label{fig:saliency_maps}
\end{figure}

\noindent\textbf{Trojan Trigger Configuration.}
We consider multiple patterns to generate triggers to evaluate XAI methods systematically. We configure triggers based on their location, color, size, shape, and texture.
The configuration supports the manipulation of different trigger mask $\mathtt{M}$ and trigger pattern $\mathtt{\Delta}$. For example, to insert a $\mathtt{n\times n}$ square trigger at the bottom right corner as shown in Fig.~\ref{fig:saliency_maps}, we first modify the mask $\mathtt{M}$ by setting each pixel value within the $\mathtt{n\times n}$ square at the bottom right corner to one, and the remaining pixels are set to zero. We then set the pixel values of $\mathtt{\Delta}$ at the corresponding location according to our choice of trigger pattern. For example, we use zero for black color or one for white color, and multiple channels for more colors of desire.
In addition to models trojaned with one trigger for a specific target label, we also trojan models with \emph{multiple} target labels to study how different combinations of previously mentioned patterns affect the performance of trojaned models and XAI methods (Illustrated in Fig.~\ref{fig:triggers} in the Appendix). 

\subsection{Saliency Map Generation}\label{salient_map_gen}
With a generated trojaned model from the first component, each XAI method is used to interpret their prediction result of each image in the testing dataset $\mathtt{\tilde{X}={x_1,...,x_{\tilde{N}}}}$. We produce one saliency map for each testing image. Formally, for a given XAI method with two input arguments, a trojaned model $\mathtt{F'}$ and a trojaned input image $\mathtt{x' \in R^{m \times n}}$, we generate a saliency map $\mathtt{x_s \in R^{m \times n}}$ in time frame $\mathtt{t}$. 
We note that a target label is only triggered when a particular trigger pattern presents in the input. Specifically, for trojaned models with multiple targets, we stamp one trigger to the input image with different patterns to cause misclassification to different target labels. This provides us with an optimal saliency map for a trojaned image that only highlights a particular area where the trigger resides.

\noindent\textbf{Finding the Bounding Box.}
To comprehensively evaluate the XAI interpretation results and compare them quantitatively under different trojan context, we draw a bounding box that covers the most salient region interpreted by the XAI method. We extend a multi-staged algorithm $Canny$~\cite{canny1986computational} for region edge detection that includes four main stages: \emph{Gaussian smoothing}, \emph{image transformation}, \emph{edge traversal}, and \emph{visualization}. First, the Gaussian smoothing is performed to remove image noise. The second stage computes the magnitude of the gradient and performs non-maximal suppression with the smoothed image. Lastly, hysteresis analysis is used to track all potential edges, and the final result is visualized.
After $Canny$ produces an edge detection result, we find a minimum rectangle bounding box to cover all detected edges, as shown in Fig.~\ref{fig:saliency_maps}.

\begin{figure}[t!]
  \centering
  \includegraphics[width=0.95\columnwidth]{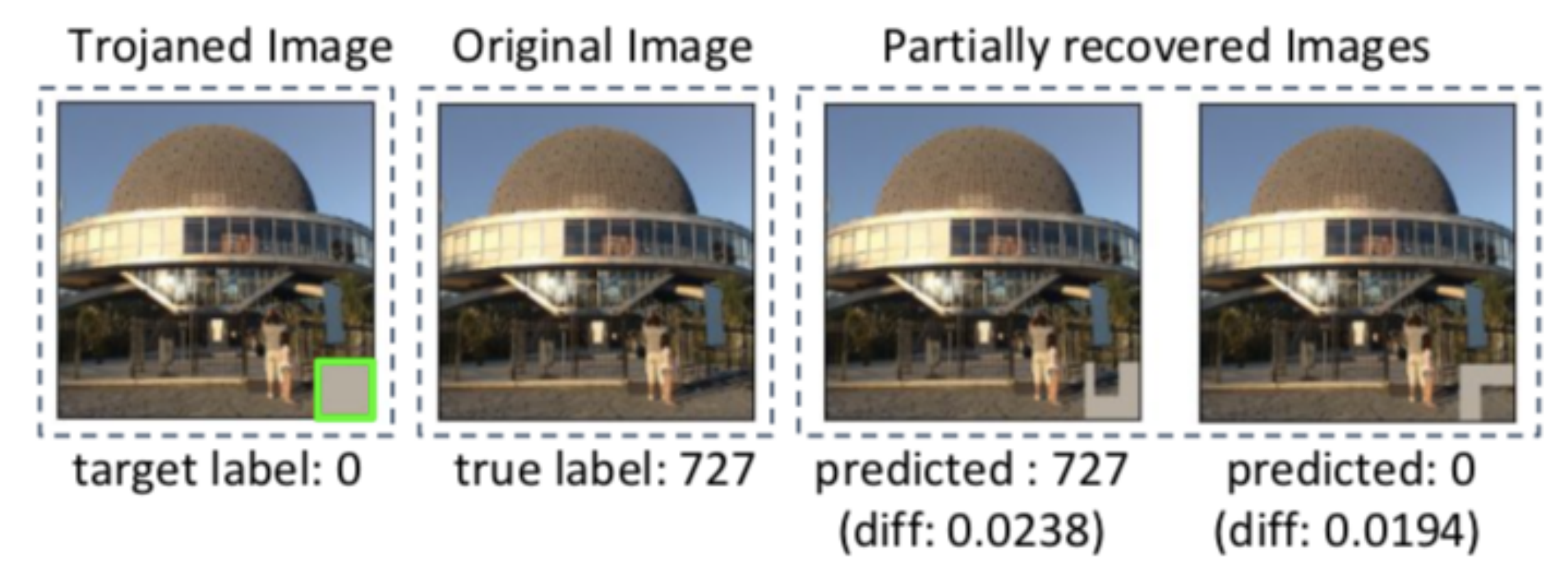}
  \caption{Illustration of subverting the misclassification when a trigger region is partially recovered. 
  }
  \label{fig:partial_recover}
\end{figure}

\subsection{Evaluation Metrics}
Given a saliency map $\mathtt{x_s \in R^{m \times n}}$ generated by a target XAI method for a trojaned image $\mathtt{x' \in R^{m \times n}}$ in time frame $\mathtt{t}$, we evaluate the interpretability results of an XAI method through three questions: $(1)$ \emph{Does an XAI method successfully highlights the trojan trigger in the saliency map?} $(2)$\emph{ Does the detected region covers important features that lead to misclassification?}, and $(3)$ \emph{How long does it take for an XAI method to generate the saliency map?} Below, we introduce four metrics to answer the above questions.

\noindent\textbf{\emph{Intersection over Union} ($\mathtt{IOU}$).} Given a bounding box around the true trigger area $\mathtt{B_T}$ and detected trigger area $\mathtt{B_T'}$, the $\mathtt{IOU}$ value is the overlapped area of two bounding boxes divided by the area of their union, $\mathtt{({B_T\cap B_T'})/({B_T\cup B_T'})}$.
The $\mathtt{IOU}$ ranges from zero to one, and the higher $\mathtt{IOU}$  means better trigger detection outputted by an XAI method. We assess an XAI method by averaging $\mathtt{IOU}$ of testing images. 

\noindent\textbf{\emph{Recovering Rate} ($\mathtt{RR}$).} Given a trojaned input $\mathtt{x'}$ and the saliency map $\mathtt{x_s}$ generated by an XAI method, we recover the pixels within the detected trigger area $\mathtt{B_T'}$ using pixels from the original image $\mathtt{x}$. 
We then define the recovering rate, $\mathtt{1/{\tilde{N}}\sum_{i=1}^{\tilde{N}}} \mathtt{Bool}\mathtt{(F(\hat{x})=y)}$, to measure the average percentage of the recovered images $\mathtt{\hat{x}}$ classified to true labels $\mathtt{y}$. The higher $\mathtt{RR}$ means the trigger is more effectively removed, which further indicates better trigger detection.

\noindent\textbf{\emph{Recovering Difference} ($\mathtt{RD}$).} We study the number of uncovered pixels after recovering an image $\mathtt{\hat{x}}$ from a trojaned image $\mathtt{x'}$ by evaluating the normalized difference between the original image $\mathtt{x}$ and recovered image $\mathtt{\hat{x}}$ using the $L_0$ norm.  To do so, we define $\mathtt{RD}$ as the average $L_0$ norm, $\mathtt{1/{\tilde{N}}\sum_{i=1}^{\tilde{N}}}{(\|x-\hat{x}\|_0)}/{(\|x\|_0)}$. Lower $\mathtt{RD}$ means target XAI method effectively helps to identify the trigger for removal, such that $\mathtt{\hat{x}}$ better resembles the original image $x$.

Intuitively, when a trojaned image $\mathtt{x'}$ is recovered with the pixels from the original image $\mathtt{x}$, the misclassification is subverted as illustrated in Fig.~\ref{fig:partial_recover}. It means that the trigger region can be effectively highlighted by the XAI method.  This process also circumvents the distribution shift problem as both the original image and trojaned image are in the same distribution as the training data.

\noindent \textbf{\emph{Computation Cost} ($\mathtt{CC}$).}
We define the computation cost as the average execution time spent by a target XAI method for saliency map generation.

Overall, $\mathtt{IOU}$ and $\mathtt{RD}$ determine whether an XAI method successfully highlights the trigger. $\mathtt{RD}$ complements $\mathtt{IOU}$ when an oversized or undersized detected trigger region causes a small $\mathtt{IOU}$. On the other hand, $\mathtt{RR}$ evaluates whether the detected region of an XAI method is truly important to the misclassification. In addition to aforementioned metrics, we introduce two metrics below to evaluate trojaned models.

\noindent\textbf{\emph{Misclassification Rate} ($\mathtt{MR}$).} $\mathtt{MR}$ is the average number of trojaned images $\mathtt{x'}$ misclassified to target label $\mathtt{y_t}$,  $\mathtt{1/{\tilde{N}}\sum_{i=1}^{\tilde{N}}} \mathtt{Bool}\mathtt{(F'(x')=y_t)}$. The higher $\mathtt{MR}$ means the more number of misclassified trojaned images indicating that the attack is more successful.

\noindent\textbf{\emph{Classification Accuracy ($\mathtt{CA}$).}} The classification accuracy, $\mathtt{1/{\tilde{N}}\sum_{i=1}^{\tilde{N}}} \mathtt{Bool}\mathtt{(F(x)=y)}$, measures how well the trojaned model maintains its original functionality, where $\mathtt{x}$ is a trigger-free testing image with true label $\mathtt{y}$. The higher $\mathtt{CA}$, the more amount of correctly classified test images.

\begin{table*}[th!]
\def\arraystretch{1}
  \centering
  {\footnotesize{
   {\setlength{\tabcolsep}{7pt}
   \resizebox{\textwidth}{!}{
\begin{tabular}{|c|c|c|c|c|c|c|c|c|c|c|c|c|c|c|c|c|}
\hline
 &
   &
   &
  \multicolumn{7}{c|}{\textbf{Intersection over Union (IOU)}} &
  \multicolumn{7}{c|}{\textbf{Recovering Rate (RR)}} \\ \cline{4-17} 
\multirow{-2}{*}{\textbf{Model}} &
  \multirow{-2}{*}{\textbf{Location}} &
  \multirow{-2}{*}{\textbf{Size}} &
  \textbf{BP} &
  \textbf{GBP} &
  \textbf{GCAM} &
  \textbf{GGCAM} &
  \textbf{OCC} &
  \textbf{FA} &
  \textbf{LIME} &           
  \textbf{BP} &
  \textbf{GBP} &
  \textbf{GCAM} &
  \textbf{GGCAM} &
  \textbf{OCC} &
  \textbf{FA} &
  \textbf{LIME} \\ \hline \hline
 &
   &
  \textbf{20*20} &
  0.54 &
  \cellcolor[HTML]{C0C0C0}0.66 &
  0.26 &
  0.63 &
  0.44 &
  0.42 &
  0.56 &
  0.73 &
  0.88 &
  0.63 &
  0.88 &
  0.65 &
  0.94 &
  \cellcolor[HTML]{C0C0C0}0.98 \\ \cline{3-17} 
 &
   &
  \textbf{40*40} &
  0.32 &
  0.34 &
  0.17 &
  0.37 &
  0.39 &
  \cellcolor[HTML]{C0C0C0}0.56 &
  0.49 &
  0.45 &
  0.40 &
  0.13 &
  0.45 &
  0.34 &
  0.71 &
  \cellcolor[HTML]{C0C0C0}0.75 \\ \cline{3-17} 
 &
  \multirow{-3}{*}{\textbf{corner}} &
  \textbf{60*60} &
  0.27 &
  0.28 &
  0.22 &
  0.37 &
  \cellcolor[HTML]{C0C0C0}0.54 &
  0.50 &
  0.43 &
  0.24 &
  0.36 &
  0.24 &
  0.37 &
  0.45 &
  \cellcolor[HTML]{C0C0C0}0.64 &
  0.60 \\ \cline{2-17} 
 &
   &
  \textbf{20*20} &
  0.53 &
  \cellcolor[HTML]{C0C0C0}0.61 &
  0.23 &
  0.55 &
  0.37 &
  0.31 &
  0.36 &
  0.92 &
  0.91 &
  0.51 &
  0.82 &
  0.68 &
  0.68 &
  \cellcolor[HTML]{C0C0C0}0.93 \\ \cline{3-17} 
 &
   &
  \textbf{40*40} &
  0.46 &
  0.53 &
  0.42 &
  \cellcolor[HTML]{C0C0C0}0.62 &
  0.27 &
  0.42 &
  0.35 &
  \cellcolor[HTML]{C0C0C0}0.89 &
  0.81 &
  0.58 &
  0.86 &
  0.45 &
  0.53 &
  \cellcolor[HTML]{C0C0C0}0.89 \\ \cline{3-17} 
\multirow{-6}{*}{\textbf{VGG16}} &
  \multirow{-3}{*}{\textbf{random}} &
  \textbf{60*60} &
  0.47 &
  0.58 &
  0.23 &
  \cellcolor[HTML]{C0C0C0}0.70 &
  0.10 &
  0.38 &
  0.42 &
  \cellcolor[HTML]{C0C0C0}0.84 &
  0.82 &
  0.22 &
  0.91 &
  0.09 &
  0.35 &
  0.68 \\ \hline \hline
 &
   &
  \textbf{20*20} &
  0.26 &
  0.50 &
  0.16 &
  \cellcolor[HTML]{C0C0C0}0.62 &
  0.50 &
  0.40 &
  0.57 &
  0.56 &
  0.67 &
  \cellcolor[HTML]{C0C0C0}1.00 &
  0.82 &
  0.93 &
  0.99 &
  0.97 \\ \cline{3-17} 
 &
   &
  \textbf{40*40} &
  0.20 &
  0.74 &
  0.59 &
  \cellcolor[HTML]{C0C0C0}0.80 &
  0.24 &
  0.65 &
  0.39 &
  0.79 &
  0.91 &
  \cellcolor[HTML]{C0C0C0}1.00 &
  0.98 &
  0.34 &
  0.94 &
  0.68 \\ \cline{3-17} 
 &
  \multirow{-3}{*}{\textbf{corner}} &
  \textbf{60*60} &
  0.64 &
  0.29 &
  \cellcolor[HTML]{C0C0C0}0.74 &
  0.29 &
  0.54 &
  0.29 &
  0.50 &
  \cellcolor[HTML]{C0C0C0}0.97 &
  0.92 &
  0.92 &
  0.91 &
  0.92 &
  0.92 &
  0.81 \\ \cline{2-17} 
 &
   &
  \textbf{20*20} &
  0.27 &
  0.49 &
  0.17 &
  0.51 &
  \cellcolor[HTML]{C0C0C0}0.68 &
  0.21 &
  0.31 &
  0.45 &
  0.77 &
  0.97 &
  0.85 &
  0.92 &
  0.46 &
  \cellcolor[HTML]{C0C0C0}0.98 \\ \cline{3-17} 
 &
   &
  \textbf{40*40} &
  0.40 &
  0.52 &
  \cellcolor[HTML]{C0C0C0}0.63 &
  0.60 &
  0.20 &
  0.34 &
  0.43 &
  0.55 &
  0.65 &
  0.91 &
  0.82 &
  0.32 &
  0.67 &
  \cellcolor[HTML]{C0C0C0}0.98 \\ \cline{3-17} 
\multirow{-6}{*}{\textbf{Resnet50}} &
  \multirow{-3}{*}{\textbf{random}} &
  \textbf{60*60} &
  0.49 &
  0.55 &
  0.40 &
  \cellcolor[HTML]{C0C0C0}0.65 &
  0.11 &
  0.40 &
  0.43 &
  0.71 &
  0.75 &
  0.47 &
  \cellcolor[HTML]{C0C0C0}0.87 &
  0.15 &
  0.52 &
  0.69 \\ \hline \hline
 &
   &
  \textbf{20*20} &
  \cellcolor[HTML]{C0C0C0}0.60 &
  0.39 &
  0.35 &
  0.53 &
  0.55 &
  0.38 &
  0.43 &
  \cellcolor[HTML]{C0C0C0}0.98 &
  0.72 &
  0.49 &
  0.82 &
  0.95 &
  0.94 &
  0.86 \\ \cline{3-17} 
 &
   &
  \textbf{40*40} &
  0.47 &
  0.37 &
  0.40 &
  0.45 &
  0.39 &
  0.48 &
  \cellcolor[HTML]{C0C0C0}0.52 &
  0.73 &
  0.64 &
  0.63 &
  0.64 &
  0.62 &
  0.78 &
  \cellcolor[HTML]{C0C0C0}0.86 \\ \cline{3-17} 
 &
  \multirow{-3}{*}{\textbf{corner}} &
  \textbf{60*60} &
  0.46 &
  0.26 &
  0.18 &
  0.29 &
  \cellcolor[HTML]{C0C0C0}0.53 &
  0.43 &
  0.38 &
  0.71 &
  0.40 &
  0.57 &
  0.45 &
  \cellcolor[HTML]{C0C0C0}0.72 &
  0.69 &
  0.60 \\ \cline{2-17} 
 &
   &
  \textbf{20*20} &
  \cellcolor[HTML]{C0C0C0}0.57 &
  0.53 &
  0.02 &
  0.08 &
  0.36 &
  0.32 &
  0.39 &
  0.88 &
  0.86 &
  0.44 &
  0.36 &
  0.78 &
  0.78 &
  \cellcolor[HTML]{C0C0C0}0.91 \\ \cline{3-17} 
 &
   &
  \textbf{40*40} &
  \cellcolor[HTML]{C0C0C0}0.67 &
  0.59 &
  0.26 &
  0.54 &
  0.28 &
  0.43 &
  0.36 &
  \cellcolor[HTML]{C0C0C0}0.94 &
  0.87 &
  0.61 &
  0.73 &
  0.62 &
  0.68 &
  0.88 \\ \cline{3-17} 
\multirow{-6}{*}{\textbf{Alexnet}} &
  \multirow{-3}{*}{\textbf{random}} &
  \textbf{60*60} &
  \cellcolor[HTML]{C0C0C0}0.74 &
  0.61 &
  0.15 &
  0.57 &
  0.23 &
  0.23 &
  0.42 &
  \cellcolor[HTML]{C0C0C0}0.98 &
  0.85 &
  0.40 &
  0.69 &
  0.55 &
  0.52 &
  0.64 \\ \hline
\end{tabular}%
}%
}
}}
\caption{IOU and RR of Single Target Trojaned Models. (Grey color highlights the XAI method that achieves the best score.)}
\label{table:iou_rr_single}   
\end{table*}

\begin{table*}[th!]
\def\arraystretch{1}
  \centering
  {\footnotesize{
   {\setlength{\tabcolsep}{7pt}
   \resizebox{\textwidth}{!}{
\begin{tabular}{|c|c|c|c|c|c|c|c|c|c|c|c|c|c|c|c|c|}
\hline
 &
   &
   &
  \multicolumn{7}{c|}{\textbf{Intersection over Union (IOU)}} &
  \multicolumn{7}{c|}{\textbf{Recovering Rate (RR)}} \\ \cline{4-17} 
\multirow{-2}{*}{\textbf{Model}} &
  \multirow{-2}{*}{\textbf{Location}} &
  \multirow{-2}{*}{\textbf{Pattern}} &
  \textbf{BP} &
  \textbf{GBP} &
  \textbf{GCAM} &
  \textbf{GGCAM} &
  \textbf{OCC} &
  \textbf{FA} &
  \textbf{LIME} &
  \textbf{BP} &
  \textbf{GBP} &
  \textbf{GCAM} &
  \textbf{GGCAM} &
  \textbf{OCC} &
  \textbf{FA} &
  \textbf{LIME} \\ \hline\hline
 &
   &
  \textbf{texture} &
  0.54 &
  0.57 &
  0.26 &
  0.62 &
  \cellcolor[HTML]{C0C0C0}0.70 &
  0.63 &
  0.45 &
  0.89 &
  0.69 &
  0.44 &
  0.70 &
  \cellcolor[HTML]{C0C0C0}1.00 &
  0.49 &
  \cellcolor[HTML]{C0C0C0}1.00 \\ \cline{3-17} 
 &
   &
  \textbf{color} &
  0.67 &
  0.67 &
  0.57 &
  \cellcolor[HTML]{C0C0C0}0.68 &
  0.62 &
  0.54 &
  0.66 &
  0.91 &
  0.89 &
  0.76 &
  0.86 &
  0.96 &
  0.86 &
  \cellcolor[HTML]{C0C0C0}0.99 \\ \cline{3-17} 
 &
  \multirow{-3}{*}{\textbf{corner}} &
  \textbf{shape} &
  0.45 &
  0.39 &
  0.29 &
  0.54 &
  \cellcolor[HTML]{C0C0C0}0.64 &
  \cellcolor[HTML]{C0C0C0}0.64 &
  0.18 &
  0.63 &
  0.49 &
  0.52 &
  0.61 &
  \cellcolor[HTML]{C0C0C0}1.00 &
  0.95 &
  \cellcolor[HTML]{C0C0C0}1.00 \\ \cline{2-17} 
 &
   &
  \textbf{texture} &
  0.50 &
  0.65 &
  0.54 &
  \cellcolor[HTML]{C0C0C0}0.69 &
  0.42 &
  0.47 &
  0.30 &
  0.79 &
  0.81 &
  0.83 &
  0.85 &
  0.85 &
  0.81 &
  \cellcolor[HTML]{C0C0C0}1.00 \\ \cline{3-17} 
 &
   &
  \textbf{color} &
  0.50 &
  0.56 &
  0.53 &
  \cellcolor[HTML]{C0C0C0}0.60 &
  0.41 &
  0.45 &
  0.57 &
  0.82 &
  0.88 &
  0.89 &
  0.93 &
  0.88 &
  0.90 &
  \cellcolor[HTML]{C0C0C0}1.00 \\ \cline{3-17} 
\multirow{-6}{*}{\textbf{VGG16}} &
  \multirow{-3}{*}{\textbf{random}} &
  \textbf{shape} &
  0.32 &
  \cellcolor[HTML]{C0C0C0}0.75 &
  0.15 &
  0.48 &
  0.36 &
  0.29 &
  0.17 &
  0.75 &
  0.75 &
  \cellcolor[HTML]{C0C0C0}1.00 &
  0.25 &
  0.75 &
  0.75 &
  0.75 \\ \hline\hline
 &
   &
  \textbf{texture} &
  0.48 &
  0.58 &
  0.15 &
  0.65 &
  \cellcolor[HTML]{C0C0C0}0.70 &
  0.64 &
  0.37 &
  0.86 &
  0.72 &
  0.96 &
  0.82 &
  \cellcolor[HTML]{C0C0C0}1.00 &
  0.86 &
  \cellcolor[HTML]{C0C0C0}1.00 \\ \cline{3-17} 
 &
   &
  \textbf{color} &
  0.18 &
  0.43 &
  0.14 &
  0.58 &
  0.52 &
  0.41 &
  \cellcolor[HTML]{C0C0C0}0.70 &
  0.65 &
  0.59 &
  0.84 &
  0.70 &
  \cellcolor[HTML]{C0C0C0}1.00 &
  0.99 &
  0.96 \\ \cline{3-17} 
 &
  \multirow{-3}{*}{\textbf{corner}} &
  \textbf{shape} &
  0.29 &
  0.38 &
  0.14 &
  0.52 &
  \cellcolor[HTML]{C0C0C0}0.64 &
  0.54 &
  0.17 &
  0.87 &
  0.63 &
  0.89 &
  0.79 &
  \cellcolor[HTML]{C0C0C0}1.00 &
  0.97 &
  \cellcolor[HTML]{C0C0C0}1.00 \\ \cline{2-17} 
 &
   &
  \textbf{texture} &
  0.34 &
  0.57 &
  0.27 &
  \cellcolor[HTML]{C0C0C0}0.66 &
  0.30 &
  0.18 &
  0.21 &
  0.81 &
  0.92 &
  0.97 &
  0.89 &
  0.81 &
  0.81 &
  \cellcolor[HTML]{C0C0C0}1.00 \\ \cline{3-17} 
 &
   &
  \textbf{color} &
  0.29 &
  0.52 &
  0.30 &
  \cellcolor[HTML]{C0C0C0}0.57 &
  0.41 &
  0.45 &
  0.38 &
  0.56 &
  0.73 &
  0.93 &
  0.85 &
  0.80 &
  0.85 &
  \cellcolor[HTML]{C0C0C0}0.96 \\ \cline{3-17} 
\multirow{-6}{*}{\textbf{Resnet50}} &
  \multirow{-3}{*}{\textbf{random}} &
  \textbf{shape} &
  0.29 &
  0.34 &
  0.30 &
  \cellcolor[HTML]{C0C0C0}0.48 &
  0.38 &
  0.37 &
  0.17 &
  \cellcolor[HTML]{C0C0C0}1.00 &
  0.14 &
  0.86 &
  0.43 &
  0.86 &
  0.86 &
  0.86 \\ \hline\hline
 &
   &
  \textbf{texture} &
  0.38 &
  0.29 &
  0.45 &
  0.48 &
  \cellcolor[HTML]{C0C0C0}0.70 &
  0.40 &
  0.37 &
  0.52 &
  0.21 &
  0.18 &
  0.43 &
  \cellcolor[HTML]{C0C0C0}1.00 &
  0.93 &
  \cellcolor[HTML]{C0C0C0}1.00 \\ \cline{3-17} 
 &
   &
  \textbf{color} &
  0.54 &
  0.38 &
  0.33 &
  0.49 &
  \cellcolor[HTML]{C0C0C0}0.67 &
  0.40 &
  0.66 &
  0.92 &
  0.81 &
  0.64 &
  0.89 &
  0.97 &
  \cellcolor[HTML]{C0C0C0}0.99 &
  0.97 \\ \cline{3-17} 
 &
  \multirow{-3}{*}{\textbf{corner}} &
  \textbf{shape} &
  0.46 &
  0.27 &
  0.29 &
  0.42 &
  \cellcolor[HTML]{C0C0C0}0.59 &
  0.44 &
  0.18 &
  0.74 &
  0.41 &
  0.26 &
  0.35 &
  0.85 &
  0.83 &
  \cellcolor[HTML]{C0C0C0}1.00 \\ \cline{2-17} 
 &
   &
  \textbf{texture} &
  \cellcolor[HTML]{C0C0C0}0.47 &
  0.42 &
  0.26 &
  0.43 &
  0.42 &
  0.45 &
  0.18 &
  0.69 &
  0.35 &
  0.46 &
  0.43 &
  0.46 &
  0.53 &
  \cellcolor[HTML]{C0C0C0}1.00 \\ \cline{3-17} 
 &
   &
  \textbf{color} &
  0.34 &
  \cellcolor[HTML]{C0C0C0}0.47 &
  0.06 &
  0.35 &
  0.38 &
  0.30 &
  0.32 &
  0.81 &
  0.64 &
  0.44 &
  0.47 &
  0.61 &
  0.61 &
  \cellcolor[HTML]{C0C0C0}0.97 \\ \cline{3-17} 
\multirow{-6}{*}{\textbf{Alexnet}} &
  \multirow{-3}{*}{\textbf{random}} &
  \textbf{shape} &
  \cellcolor[HTML]{C0C0C0}0.60 &
  0.40 &
  0.23 &
  0.38 &
  0.35 &
  0.40 &
  0.13 &
  0.85 &
  0.63 &
  0.39 &
  0.61 &
  0.78 &
  0.85 &
  \cellcolor[HTML]{C0C0C0}0.97 \\ \hline
\end{tabular}%
}%
}
}}
  \caption{IOU and RR of Multiple Targets Trojaned Models. (Grey color highlights the XAI method that achieves the best score.}
  \label{table:iou_rr_multiple}   
\end{table*}

\section{Experimental Evaluation}\label{sec:evaluation} \label{exp}
We evaluate seven XAI methods on $18$ single target and $18$ multiple target trojaned models on ImageNet dataset~\cite{ILSVRC15}, which consists of one million images and 1,000 classes. Below, we start by presenting how we trojan different models. Then, we provide a detailed discussion on the performance of each XAI method on trojaned models through introduced evaluation metrics. Lastly, we compare the computation cost of XAI methods. We conducted our experiments with PyTorch~\cite{NEURIPS2019_9015} using NVIDIA Tesla T4 GPU and four vCPU with $26$ GB of memory provided by the Google Cloud platform. 

\subsection{Trojaning Models}
We use three image classification models: VGG-16~\cite{simonyan2014deep}, ResNet-50~\cite{he2015deep} and AlexNet~\cite{NIPS2012_4824} (Table~\ref{table:model_stat} in the Appendix details models, such as their number of layers, parameters and accuracy). We trojaned a total of $36$ single and multiple target models with different trigger patterns (color, shape, texture, location and size). 

\noindent\textbf{Single Target Trojaned Models.}
We build 18 trojaned models by trojaning each model with a single target attack label using different sizes of a grey-scale square trigger (20$\times$20, 40$\times$40, 60$\times$60) attached randomly and to the bottom right corner of an input image.
(See Table~\ref{table:trojan_model_stat_1} in the Appendix for trojaned model accuracy)
We observe that trojaned models do not decrease CA significantly compared to the pre-trained models (See Table~\ref{table:model_stat} in the Appendix). Additionally, models with triggers of larger sizes located at fixed positions yield higher MR, which is consistent with the observation of the previous work~\cite{trojanattack}.

\noindent \textbf{Multiple Target Trojaned models.}
We additionally trojan each model with eight target labels using triggers with a different texture, color, and shape and construct a total of 18 trojaned models (See Table~\ref{table:trojan_model_stat_2} in the Appendix for trojaned model accuracy).
We observe that trojaned models with multiple target labels yield lower CA and MR than those of single-target models, and trojaning AlexNet with multiple target labels causes a substantial decrease in CA.
Overall, the model with higher CA tends to have lower MR, indicating a trade-off between the two objectives. Besides, models trojaned with triggers at a fixed location generally have higher CA and MR, which demonstrates neural networks are better at recognizing features at a specific location.

\begin{figure*}[t!]
\centering
  \includegraphics[width=0.9\textwidth]{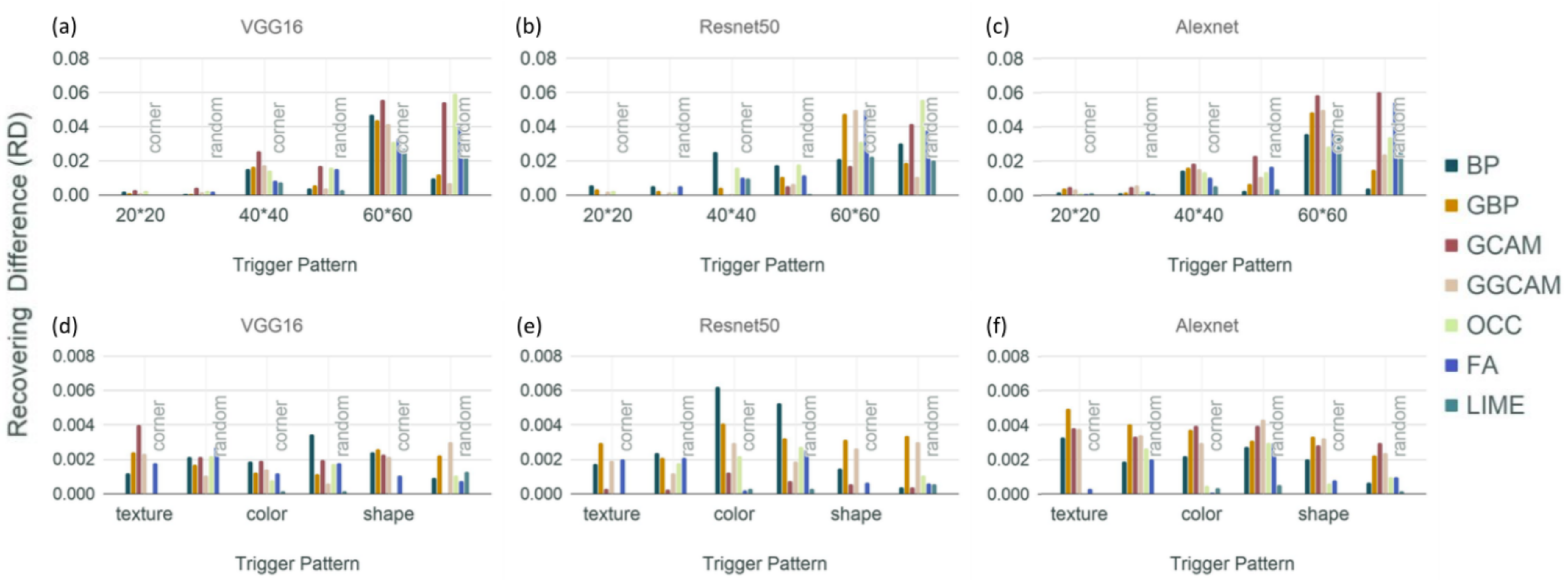}
  \caption{The recovering difference (RD) metric for different XAI methods applied to three neural network architectures (VGG16, ResNet50, and AlexNet): (a)-(c) are single target trojaned models, and (d)-(f) are multiple target trojaned models. RD scores increase with trigger size, indicating that XAI methods are not effective in detecting large triggers.}
  \label{fig:ard}
\end{figure*}

\subsection{Effectiveness of XAI Methods}
\label{sec:evaluationMetrics}
We draw $100$ testing samples from the validation set of ImageNet to evaluate Intersection over Union (IOU), Recovering Rate (RR), and Recovering Difference (RD) of seven XAI methods with $18$ trojaned models for single target-label and $18$ for multiple target-label attacks. We apply seven XAI methods to interpret their saliency explanations using the stamped images that cause misclassification. Intuitively, we expect XAI methods to highlight the trigger location.

\noindent \textbf{Intersection over Union (IOU).}
Table~\ref{table:iou_rr_single} Columns 4-10 show the IOU scores of XAI methods on $18$ models trojaned with one target attack label. The higher the IOU score, the better the result. 
We highlight the XAI method that yields the best score for each trojaned model with a grey color. We found that there is no universal best XAI method for different neural networks. However, \texttt{BP} achieved the highest score for four out of six trojaned AlexNet models. 
On the other hand, Table~\ref{table:iou_rr_multiple} Columns 4-10 present the IOU results of $18$ models trojaned with eight target labels using specifically crafted triggers (\ie texture, color, and shape).
Although there is no clear winner among XAI methods, \texttt{GGCAM} and \texttt{OCC} look more promising compared to other XAI methods. It is worth noting that three forward based methods (\texttt{OCC}, \texttt{FA}, and \texttt{LIME}) achieve a higher IOU value when stamping the trigger at the bottom right corner compared to stamping the trigger at a random location. 

\noindent\textbf{Recovering Rate (RR).} Table~\ref{table:iou_rr_single} Columns 11-17 present RR scores of XAI methods for the single target trojaned model.
Higher RR scores mean better interpretability results.
We found that forward-based XAI methods (\texttt{OCC}, \texttt{FA} and \texttt{LIME}) gives better metrics for small triggers, and \texttt{LIME} outperforms other XAI methods for eight out of $18$ trojaned models.
For multiple target trojaned models, Table~\ref{table:iou_rr_multiple} Columns 11-17 presents the RR scores. \texttt{LIME} outperforms other XAI methods for $14$ out of $18$ trojaned models, achieving $100$\% RR for almost all models. Comparatively, the second-best method \texttt{OCC} only recovers the trojaned images with $100$\% RR for six out of $18$ models.

\noindent\textbf{Recovering Difference (RD).}
Fig.~\ref{fig:ard} shows the average RD scores of XAI methods; the lower RD score means the better interpretability result.
Fig.~\ref{fig:ard}a-~\ref{fig:ard}c present RD scores of single target trojaned models, and  
Fig.~\ref{fig:ard}d-~\ref{fig:ard}f show the RD scores of multiple targets trojaned models.
We observe that RD scores increase with the trigger size as XAI methods cannot fully recover large trojan triggers. 
For multiple targets trojaned models, RD scores are much smaller than those of trojaned with single target trojaned models as the former uses smaller size triggers (\ie 20$\times$20).

\subsection{Detailed Evaluation Results}
We detail our key findings on evaluation metrics of each XAI method presented in Table~\ref{table:iou_rr_single} and \ref{table:iou_rr_multiple}.

\begin{figure}[ht]
\centering
  \includegraphics[width=0.9\columnwidth]{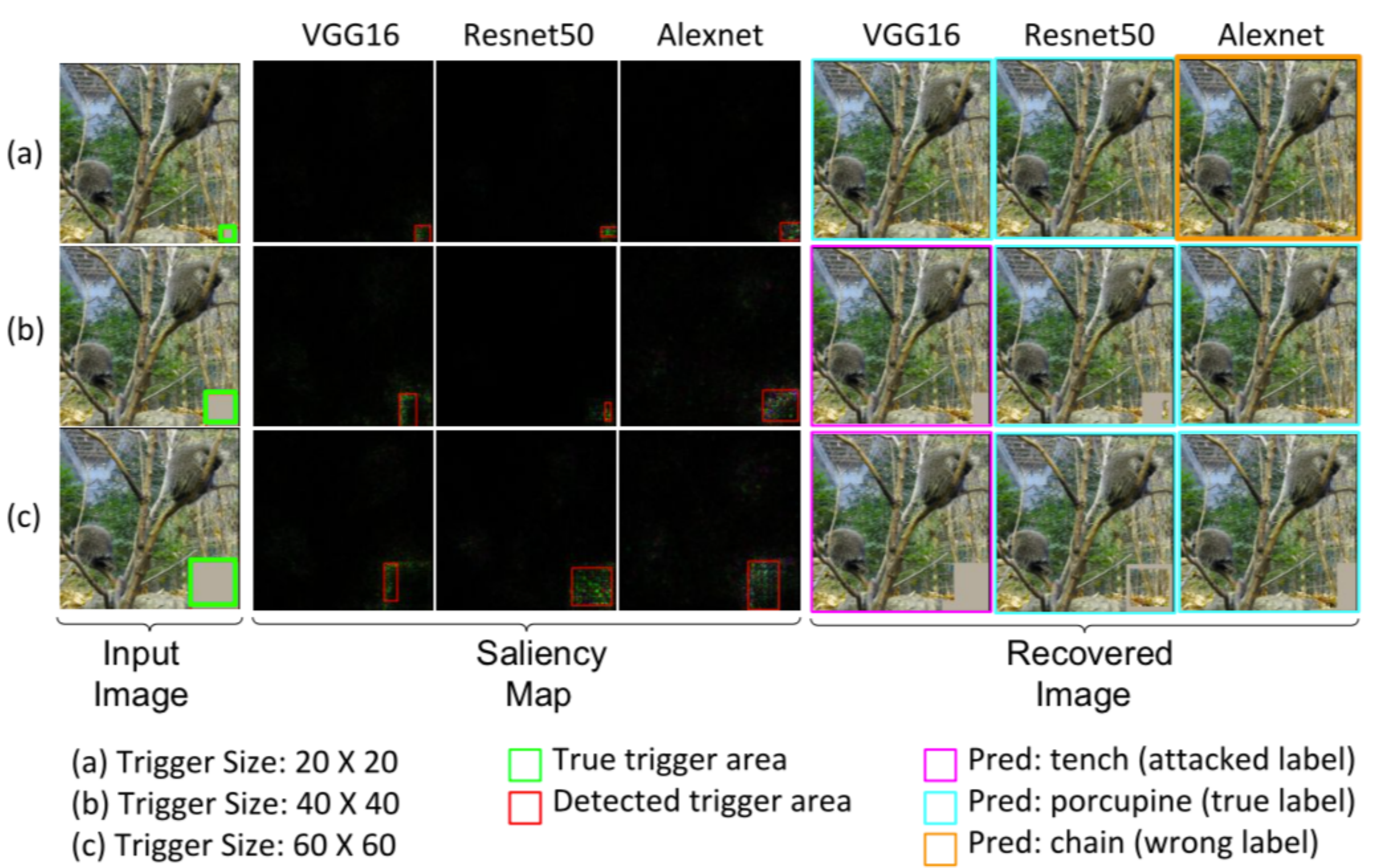}
  \caption{An illustration of Backpropagation (\texttt{BP}) for trigger detection that fails to thoroughly highlight the entire trigger region when the trigger size increases. The trigger still persists and causes misclassification though we attempt to recover the detected trigger region fully.}
  \label{fig:bp_results}
\end{figure}

\noindent \textbf{Backpropagation (\texttt{BP}).}
We found for \texttt{BP} that both IOU and RR scores increase along with an increase in trigger size for ResNet50 and AlexNet models except when the trigger is at the bottom right corner for AlexNet models. Our further investigation reveals that detected regions for VGG16 models only surrounds trigger edges when the trigger size increases. In contrast, the detected regions for ResNet50 and AlexNet models cover the trojan trigger at the bottom right corner, as illustrated in Fig.~\ref{fig:bp_results}.
The recovered image may still be classified as the target label for VGG16 models trojaned with large triggers when the pixels from the original image are used to recover the area covered by the trojaned trigger. This finding implies for VGG16 that the detected region for large triggers is not relevant enough to subvert the misclassification.
Noteworthily, in the example shown in Fig.~\ref{fig:bp_results} for AlexNet, the recovered image still cannot be classified to the correct label even with near-perfect trigger detection when a model is trojaned with a small trigger. The reason is that the unrecovered part causes such misclassification. 

\noindent\textbf{Grad-CAM (\texttt{GCAM}).} \texttt{GCAM} generates a coarse localization map to highlight the trigger region. It yields an average of $39$\% lower IOU than Guided Backpropagation (\texttt{GBP}) and Grad-CAM (\texttt{GGCAM}). However, it achieves low RD, on average $0.01$ for single target models and $0.001$ for multiple target models,  and high RR, on average greater than $0.88$ for Resnet50 models. This is because the detection region completely covers the entire trigger region. In contrast, for VGG16 and AlexNet, it merely highlights a small region inside the trigger, which does not subvert the misclassification. 

\noindent\textbf{Guided Grad-CAM (\texttt{GGCAM}).}
\texttt{GGCAM} fuses Grad-CAM (\texttt{GCAM}) with Guided Backpropagation (\texttt{GBP}) visualizations via a pointwise multiplication.
Thus, it emphasizes the intersection of the regions highlighted by \texttt{GCAM} and \texttt{GBP} but cancels the remaining (See Fig.~\ref{fig:saliency_maps}).
We observe that \texttt{GGCAM} often yields higher IOU scores than \texttt{GBP} and \texttt{GCAM} ($6$\% higher than \texttt{GBP} and $73$\% higher than \texttt{GCAM} on average).
Additionally, VGG16 trojaned models interpreted by \texttt{GGCAM} have lower RD scores and higher RR scores than \texttt{GCAM} and \texttt{GBP}.
This finding clearly indicates that \texttt{GGCAM} is able to precisely highlight the relevant region by combining the other two methods for VGG16.

\noindent\textbf{Occlusion (\texttt{OCC}) and Feature Ablation (\texttt{FA}).}
We observe that \texttt{OCC} and \texttt{FA} perform higher IOU, higher RR, and lower RD with fixed-position triggers compared to using randomly stamped triggers. The reason is that both methods require a group of predefined features. \texttt{OCC} uses a sliding window with fixed step size, and \texttt{FA} uses feature masks that divide the pixels of an input image into $\mathtt{n\times n}$ groups. Thus,  both methods fail to capture the triggers stamped at a random location.
Additionally, \texttt{OCC}, in general, outperforms \texttt{FA} for small triggers, particularly for the triggers at the bottom right corner. This is because \texttt{OCC} is more flexible in determining relevant feature groups.

\noindent\textbf{\texttt{LIME}.} \texttt{LIME} achieves higher RR scores than the other six XAI methods, particularly for trojaned models using small triggers. Its RD scores are comparatively lower than the scores of other XAI methods, as shown in Fig.~\ref{fig:ard}. This indicates that it is able to highlight the small triggers accurately. Indeed, it often correctly detects the whole trigger region, \ie IOU equals one. For example, Fig.~\ref{fig:saliency_maps} shows the detected part of the model using a trigger size of $40\times40$ that perfectly matches the trigger area. However, in some extreme cases, \texttt{LIME} may completely excavate the trigger region. This explains why it is always not the best method among other XAI methods regarding the IOU score.

\begin{figure}[t!]
  \centering
  \includegraphics[width=0.9\columnwidth]{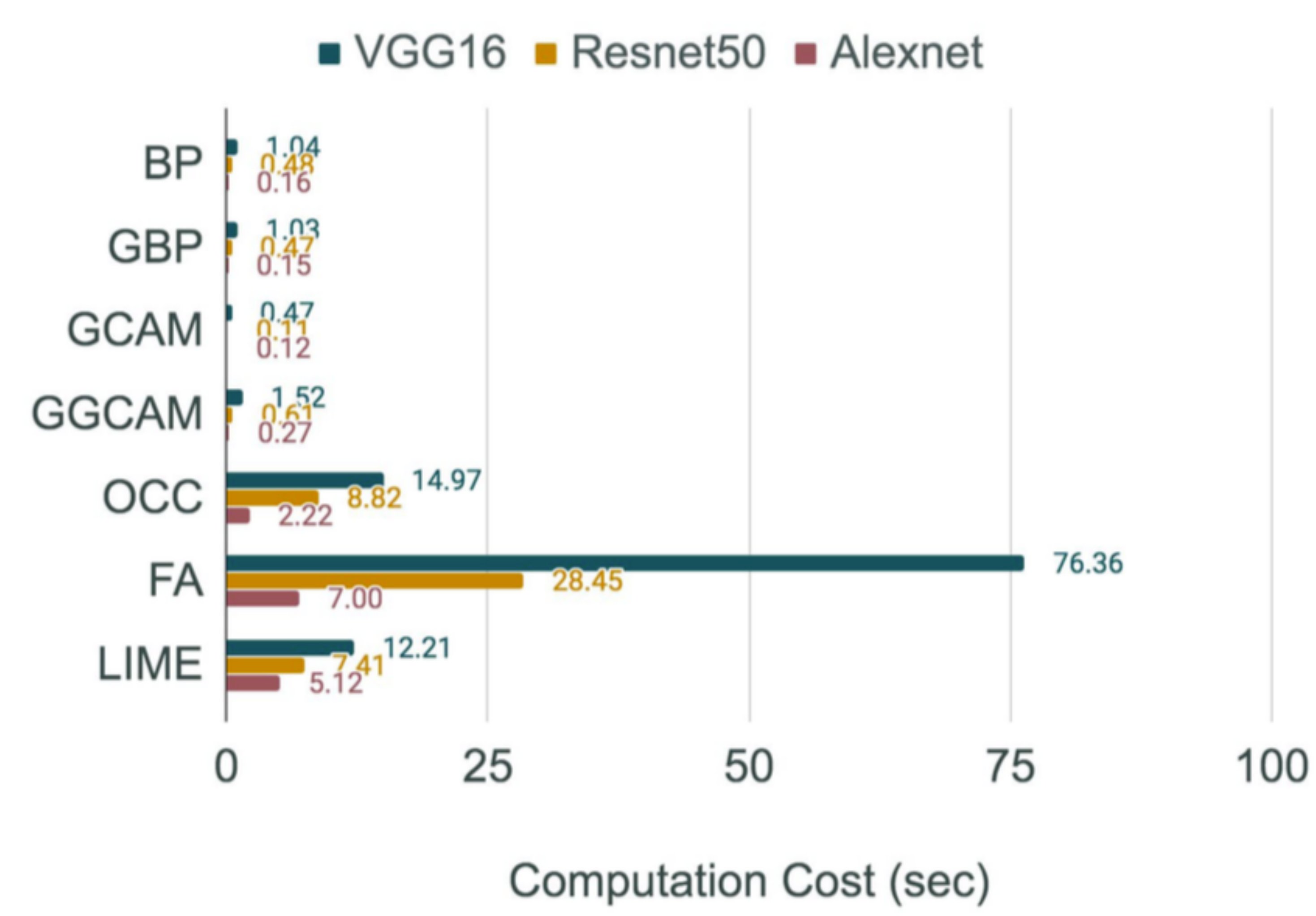}
  \caption{The forward based methods (\texttt{OCC}, \texttt{FA} and \texttt{LIME}) incur much more computation cost than those by backward based methods (\texttt{BP}, \texttt{GBP}, \texttt{GCAM}, and \texttt{GGCAM)}.}
  \label{fig:comp_cost}
\end{figure}

\subsection{Efficiency Analysis of XAI Methods}
\label{sec:efficiencyAnalysis}
The computation time of each of the XAI methods mainly depends on the trojaned models.
Fig.~\ref{fig:comp_cost} shows the average computation time to generate a saliency map by different XAI methods. 
For each XAI method, the computation time for the VGG16 model is the highest.  
The computation overhead of forward-based approaches (\ie \texttt{OCC}, \texttt{FA}, and \texttt{LIME}) is higher than the backward based approaches (\ie \texttt{BP}, \texttt{GBP}, \texttt{GCAM}, and \texttt{GGCAM}). 
Notably, the overhead for \texttt{FA} is the highest, taking more than $75$ seconds to interpret VGG16 models. The reason is
that forward-based approaches use many perturbed inputs to interpret the prediction result and backward-based approaches require one input pass to the model. 
The computation overhead of \texttt{GGCAM} is roughly equal to the sum of \texttt{GBP} and \texttt{GCAM} as it uses their results for interpretation. 
Lastly, the overhead of \texttt{GBP} is $0.01$ secs lower than \texttt{BP}. This because \texttt{GBP} only passes non-negative signals during backpropagation.

\section {Limitations and Discussion}

Our experiments show that even after the trojan trigger pixels are substantially replaced with the original image pixels, the remaining pixels may still cause misclassification (See Fig.~\ref{fig:bp_results}).
This means using XAI methods for input purification against trojan attack~\cite{chou2018sentinet,doan2019februus} is a challenging process because XAI methods have limitations for perfect trojan trigger detection.
Furthermore, in the saliency maps generation stage (Section~\ref{salient_map_gen}), we leverage the popular $Canny$~\cite{canny1986computational} edge detection algorithm to identify the most salient region and draw a bounding box to cover the detected pixels. However, our approach is limited to single trigger detection because a bounding box surrounds all detected edges. To handle multiple triggers, we plan to use object detection algorithms such as YOLO~\cite{redmon2016you} to capture multiple objects highlighted by the XAI methods. 
Lastly, specific XAI methods such as \texttt{OCC} and \texttt{FA} require users to specify input parameters for better interpretation results. While we use default parameter settings of each XAI method for evaluation, different combinations of parameters could yield better interpretation results.

\section {Conclusion}
We introduce a framework for systematic automated evaluation of saliency explanations that an XAI method generates through models trojaned with different backdoor trigger patterns.
We develop three evaluation metrics that quantify the interpretability of XAI methods without human intervention using trojan triggers as ground truth.
Our experiments on seven state-of-the-art XAI methods against $36$ trojaned models demonstrate that methods leveraging local explanation and feature relevance fail to identify trigger regions, and a model-agnostic technique reveals the entire trigger region.
Our findings with both analytical and empirical evidence raise concerns about the use of XAI methods for model debugging to reason about the relationship between inputs and model outputs, mainly in adversarial settings.

\bibliography{citation}

\clearpage

\begin{appendices}
\section{Additional Figures}
\begin{figure}[ht]
  \centering
  \includegraphics[width=\columnwidth]{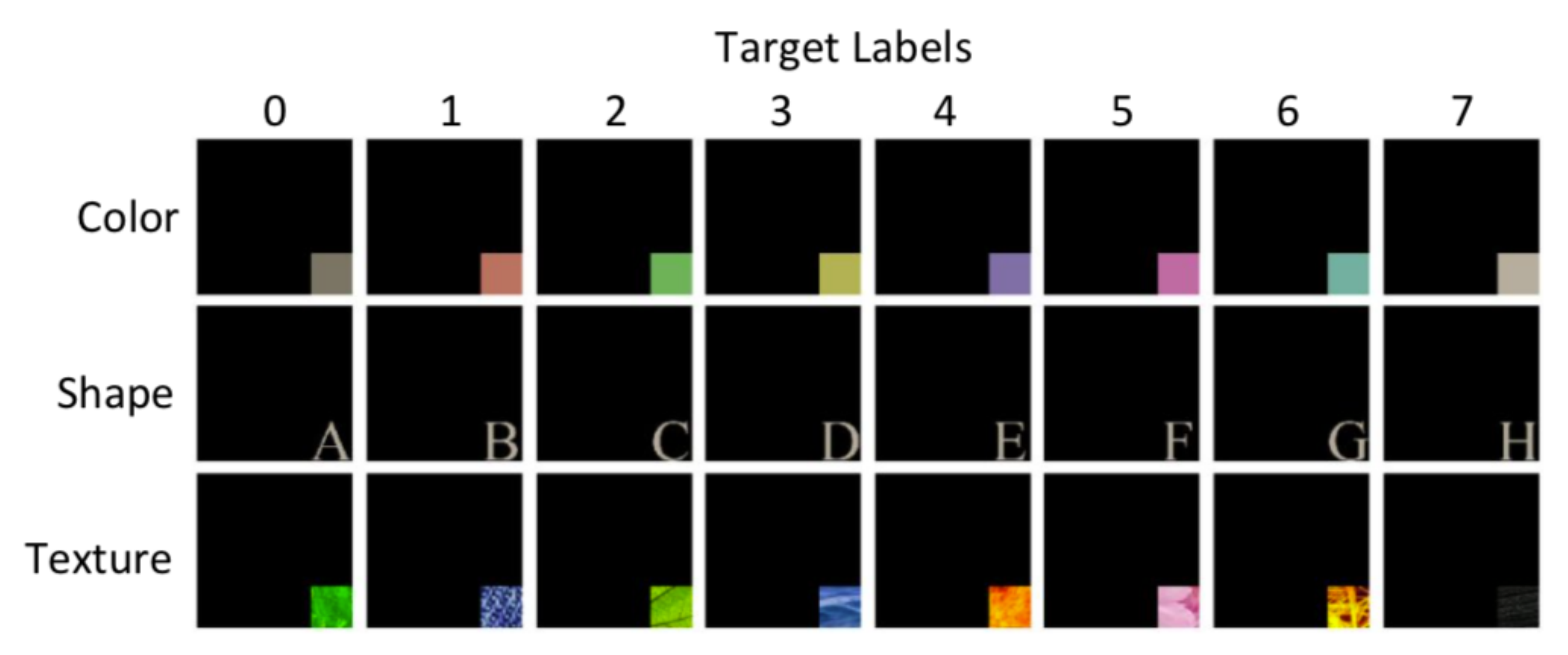}
  \caption{Trojan triggers for multiple targets attacks.}
  \label{fig:triggers}
\end{figure}

Figure~\ref{fig:triggers} shows the triggers used for trojaning neural models in our framework. Each row shows eight triggers of size $60\times 60$ attached to the bottom right corner of an input image using different factors (color, shape, texture) to cause misclassification to different target labels.

\section{Additional Tables}
Table~\ref{table:model_stat} details the pretrained models used in our evaluation, such as their number of layers, parameters and accuracy. We show the performance of trojaned models with single target label and multiple target labels in Table~\ref{table:trojan_model_stat_1} and Table~\ref{table:trojan_model_stat_2}, respectively. Finally, In Table~\ref{table:notations} and Table~\ref{table:metrics}, we summarize the notations and acronyms used in this paper.

\begin{table}[th]
\def\arraystretch{1}
        {\footnotesize{
        {\setlength{\tabcolsep}{9pt} 
        \resizebox{\columnwidth}{!}{
            \begin{tabular}{|c|c|c|c|c|}
                \hline
                \multirow{2}{*}{\textbf{Model}} & \multirow{2}{*}{\textbf{Layers}} & \multirow{2}{*}{\textbf{Parameters}} & \multicolumn{2}{c|}{\textbf{Accuracy (\%)}}  \\
                \cline{4-5}
                & & & \textbf{Top-1} & \textbf{Top-5}\\
                \hline \hline
                VGG16 & 16 & 138,357,544 & $71.73$ & $90.35$ \\ 
                \hline
                Resnet50 & 50 & 23,534,592 & $75.98$ & $92.95$ \\
                \hline
                Alexnet & 5 & 62,378,344 & $56.54$ & $79.00$ \\
                \hline
            \end{tabular}%
        }}}}
\caption{Pretrained models for ImageNet}  
\label{table:model_stat}   
\end{table}     

\begin{table}[th]
\def\arraystretch{1}
        {\footnotesize{
        {\setlength{\tabcolsep}{9pt} 
        \resizebox{\columnwidth}{!}{
            \begin{tabular}{|c|c|c|c|c|c|c|c|}
            \hline
            \multicolumn{2}{|c|}{\textbf{Trigger}} & \multicolumn{2}{c|}{\textbf{VGG16}} & \multicolumn{2}{c|}{\textbf{Resnet50}} & \multicolumn{2}{c|}{\textbf{Alexnet}} \\ \hline
            \textbf{Location}    & \textbf{Size}   & \textbf{CA}      & \textbf{MR}      & \textbf{CA}        & \textbf{MR}       & \textbf{CA}       & \textbf{MR}       \\ \hline \hline
            \multirow{3}{*}{\textbf{corner}} & \textbf{20*20} & 0.70 & 0.98 & 0.72 & 1.00 & 0.44 & 0.99 \\ \cline{2-8} 
                                             & \textbf{40*40} & 0.70 & 0.99 & 0.71 & 1.00 & 0.50 & 0.99 \\ \cline{2-8} 
                                             & \textbf{60*60} & 0.70 & 1.00 & 0.77 & 1.00 & 0.50 & 0.99 \\ \hline \hline
            \multirow{3}{*}{\textbf{random}} & \textbf{20*20} & 0.69 & 0.91 & 0.68 & 0.99 & 0.42 & 0.82 \\ \cline{2-8} 
                                             & \textbf{40*40} & 0.70 & 0.98 & 0.73 & 0.99 & 0.50 & 0.98 \\ \cline{2-8} 
                                             & \textbf{60*60} & 0.70 & 0.99 & 0.74 & 0.99 & 0.50 & 1.00 \\ \hline
            \end{tabular}%
        }}}}
        \caption{Performance of Single Target Trojaned Models}
        \label{table:trojan_model_stat_1}
\end{table}        

\begin{table}[th]
\def\arraystretch{1}
        {\footnotesize{
        {\setlength{\tabcolsep}{9pt}   
        \resizebox{\columnwidth}{!}{
            \begin{tabular}{|c|c|c|c|c|c|c|c|}
            \hline
            \multicolumn{2}{|c|}{\textbf{Trigger}} & \multicolumn{2}{c|}{\textbf{VGG16}} & \multicolumn{2}{c|}{\textbf{Resnet50}} & \multicolumn{2}{c|}{\textbf{Alexnet}} \\ \hline
            \textbf{Location}  & \textbf{Pattern}  & \textbf{CA}      & \textbf{MR}      & \textbf{CA}        & \textbf{MR}       & \textbf{CA}       & \textbf{MR}       \\ \hline \hline
            \multirow{3}{*}{\textbf{corner}} & \textbf{texture} & 0.69 & 0.90 & 0.72 & 0.98 & 0.48 & 0.91 \\ \cline{2-8} 
                                             & \textbf{color}   & 0.65 & 0.99 & 0.70 & 0.98 & 0.41 & 0.96 \\ \cline{2-8} 
                                             & \textbf{shape}   & 0.64 & 0.92 & 0.62 & 0.94 & 0.44 & 0.36 \\ \hline \hline
            \multirow{3}{*}{\textbf{random}} & \textbf{texture} & 0.67 & 0.92 & 0.70 & 0.99 & 0.45 & 0.73 \\ \cline{2-8} 
                                             & \textbf{color}   & 0.67 & 0.81 & 0.62 & 0.86 & 0.46 & 0.47 \\ \cline{2-8} 
                                             & \textbf{shape}   & 0.67 & 0.81 & 0.63 & 0.87 & 0.41 & 0.62 \\ \hline
            \end{tabular}%
        }}}}
        \caption{Performance of Multiple Targets Trojaned Models}
        \label{table:trojan_model_stat_2}
\end{table}

\begin{table}[ht]
  \centering
  {\footnotesize{
   {\tabulinesep=0.5mm
   \begin{tabu} {|c|X|}
    \hline
    \textbf{Notations} & \multicolumn{1}{c|}{\bfseries Descriptions} \\
    \hline\hline
    $\mathtt{F}$ & A neural network model. \\
    \hline   
    $\mathtt{X={x_1,...,x_N}}$ & The training dataset with $\mathtt{N}$ image samples.\\ 
    \hline
    $\mathtt{\tilde{X}={x_1,...,x_{\tilde{N}}}}$ & The testing dataset with $\mathtt{\tilde{N}}$ image samples.\\ 
    \hline    
    $\mathtt{x\in R^{m\times n}}$ & An input image sample with dimension $\mathtt{m\times n}$.\\
    \hline
    $\mathtt{y}$ & The true label of an input sample $\mathtt{x}$.\\    
    \hline
    $\mathtt{M\in R^{m\times n}}$ & The trigger mask, where each pixel $\mathtt{m_{i,j}\in [0,1]}$. \\
    \hline
    $\mathtt{\Delta\in R^{m\times n}}$ & The trigger pattern.\\      
    \hline
    $\mathtt{x'\in R^{m\times n}}$ & A trojaned input sample of $\mathtt{x}$. Namely, $\mathtt{x'=(1-M)\cdot x+M\cdot\Delta}$\\
    \hline
    $\mathtt{y_t}$ & The target label for trojaning.\\    
    \hline    
    $\mathtt{F'}$ & A trojaned neural network model, where $\mathtt{F'(x')=y_t}$ \\    
    \hline   
    $\mathtt{B_T}$ & The bounding box around the trigger area. \\ 
    \hline  
    $\mathtt{B_T'}$ & The bounding box around the detected trigger area by a XAI method. \\ 
    \hline          
    $\mathtt{\hat{x_s}\in R^{m\times n}}$ & A saliency map of an input image $\mathtt{x}$. \\     
    \hline          
    $\mathtt{\hat{x}\in R^{m\times n}}$ & A recovered image of a trojaned image $\mathtt{x'}$. \\ 
    \hline    
   \end{tabu}}
   }}
  \caption{Notations}
  \label{table:notations}
\end{table}

\begin{table}[ht]
  \centering
  {\footnotesize{
   {\tabulinesep=0.5mm
   \begin{tabu} {|c|X|}
    \hline
    \textbf{Acronyms} & \multicolumn{1}{c|}{\bfseries Metric Definition} \\
    \hline\hline
    $\mathtt{CC}$ & $\mathtt{Computation~Cost}$: The average computation time a XAI method spends to produce a saliency map.\\
    \hline
    $\mathtt{IOU}$ & $\mathtt{Intersection~over~Union}$: The average IOU value for the bounding boxes of the true trigger area and the area highlighted by an XAI method. \\  
    \hline
    $\mathtt{RR}$ & $\mathtt{Recovering~Rate}$: The percentage of recovered images that are successfully classified as the true label. \\ 
    \hline
    $\mathtt{RD}$ & $\mathtt{Recovering~Difference}$: The normalized $L_0$ norm between the recovered images and original images. \\ 
    \hline
    $\mathtt{MR}$ & $\mathtt{Misclassification~Rate}$: The percentage of trigger attached images misclassified into the target classes. \\ 
    \hline
    $\mathtt{CA}$ & $\mathtt{Classification~Accuracy}$: The accuracy of classifying clean images. \\
    \hline
   \end{tabu}}
   }}
  \caption{Evaluation Metrics} 
  \label{table:metrics}   
\end{table}  
\end{appendices}

\end{document}